\documentclass[runningheads]{llncs}
\usepackage{graphicx}
\usepackage{comment}
\usepackage{amsmath,amssymb,gensymb} %
\usepackage{color}
\usepackage[dvipsnames]{xcolor}
\usepackage{inconsolata}

\usepackage{microtype}
\usepackage{graphicx}
\usepackage{xspace}

\usepackage{epsfig}
\usepackage{graphicx}
\usepackage{wrapfig}
\usepackage{subfigure}

\usepackage{textcomp}
\usepackage{stmaryrd}
\usepackage{upgreek}
\usepackage{bm}
\usepackage{cases}
\usepackage{mathtools}
\usepackage{arydshln}
\usepackage{multirow}

\usepackage{paralist}

\usepackage{cite}

\definecolor{citecolor}{HTML}{3498DB}
\definecolor{linkcolor}{HTML}{E74C3C}
\usepackage[colorlinks=true,pagebackref=true,citecolor=citecolor,linkcolor=linkcolor]{hyperref}

\usepackage{multirow}
\usepackage{rotating}
\usepackage{booktabs}
\usepackage{tabularx}
\usepackage{colortbl}

\usepackage{url}
\usepackage{xspace}
\usepackage[T1]{fontenc}

\newcolumntype{s}{>{\columncolor[gray]{.85}[.5\tabcolsep]}c}

\newcommand{\navg}[0]{nav-graph\xspace}

\newcommand{\basemodel}{Seq2Seq\xspace}

\newcommand{\xhdr}[1]{\vspace{3pt}\noindent\textbf{#1}}
\newcommand{\myquote}[1]{``\emph{#1}''}

\newcommand{\SUPP}[1]{#1}

\newcommand{\tabref}[1]{Tab.~\ref{#1}\xspace}
\newcommand{\figref}[1]{Fig.~\ref{#1}\xspace}
\newcommand{\secref}[1]{Sec.~\ref{#1}\xspace}

\newcommand{\etal}{et al.\xspace}
\newcommand{\eg}{e.g.\xspace}
\newcommand{\ie}{i.e.\xspace}
\newcommand{\vs}{vs.\xspace}

\DeclareMathOperator*{\argmax}{argmax}

\belowdisplayskip \abovedisplayskip

\newcommand{\csection}[1]{
    \vspace{-0.11in}
    \section{#1}
    \vspace{-0.10in}
}

\newcommand{\csubsection}[1]{
    \vspace{-0.09in}
    \subsection{#1}
    \vspace{-0.08in}
}

\begin{document}
\pagestyle{headings}
\mainmatter
\def\ECCVSubNumber{6018}

\title{Beyond the Nav-Graph: Vision-and-Language Navigation in Continuous Environments}
\titlerunning{Vision-and-Language Navigation in Continuous Environments}

\author{
    Jacob Krantz\inst{1} \and
    Erik Wijmans\inst{2,3} \and
    Arjun Majumdar\inst{2} \and\\
    Dhruv Batra\inst{2,3} \and
    Stefan Lee\inst{1}
}
\authorrunning{J. Krantz et al.}

\institute{Oregon State University \and
Georgia Institute of Technology \and
Facebook AI Research
\newline \href{https://jacobkrantz.github.io/vlnce}{jacobkrantz.github.io/vlnce}
}

\maketitle

\begin{abstract}
\vspace{-0.15in}

We develop a language-guided navigation task set in a continuous 3D environment where agents must execute low-level actions to follow natural language navigation directions. 
By being situated in continuous environments, this setting lifts a number of assumptions implicit in prior work that represents environments as a sparse graph of panoramas with edges corresponding to navigability. Specifically, our setting drops the presumptions of known environment topologies, short-range oracle navigation, and perfect agent localization. To contextualize this new task, we develop models that mirror many of the advances made in prior settings as well as single-modality baselines. While some of these techniques transfer, we find significantly lower absolute performance in the continuous setting -- suggesting that performance in prior `navigation-graph' settings may be inflated by the strong implicit assumptions. 

\end{abstract}
\vspace{-0.15in}
\csection{Introduction}
\label{sec:intro}

Springing forth from the pages of science fiction and capturing the daydreams of weary chore-doers everywhere, the promise and potential of general-purpose robotic assistants that follow natural language instructions has been long understood. Taking a small step towards this goal, recent work has begun developing artificial agents that follow natural language navigation instructions in perceptually-rich, simulated environments \cite{anderson2018vision,chen2019touchdown}. An example instruction might be \myquote{Go down the hall and turn left at the wooden desk. Continue until you reach the kitchen and then stop by the kettle.} and agents are evaluated by their ability to follow the described path in (potentially novel) simulated environments.

\begin{figure}[t]
    \centering
    \subfigure[\scriptsize Vision-and-Language Navigation (VLN)\label{fig:vln}]{
        \includegraphics[width=0.49\textwidth]{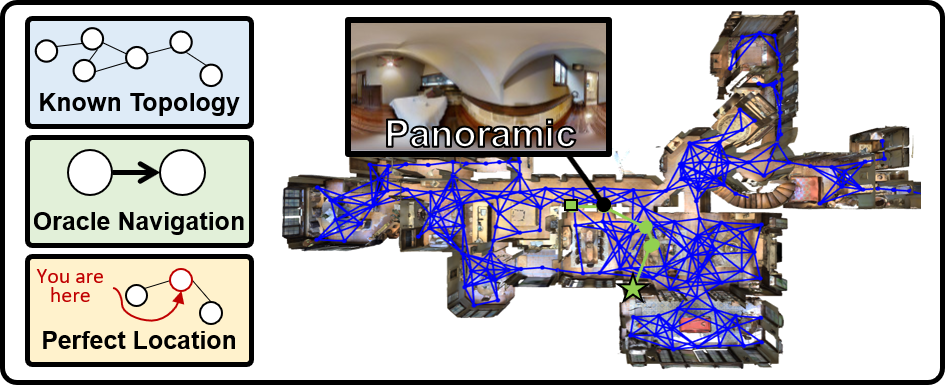}}\hfill
    \subfigure[\scriptsize VLN in Continuous Environments (VLN-CE) \label{fig:vlnotr}]{
        \includegraphics[width=0.49\textwidth]{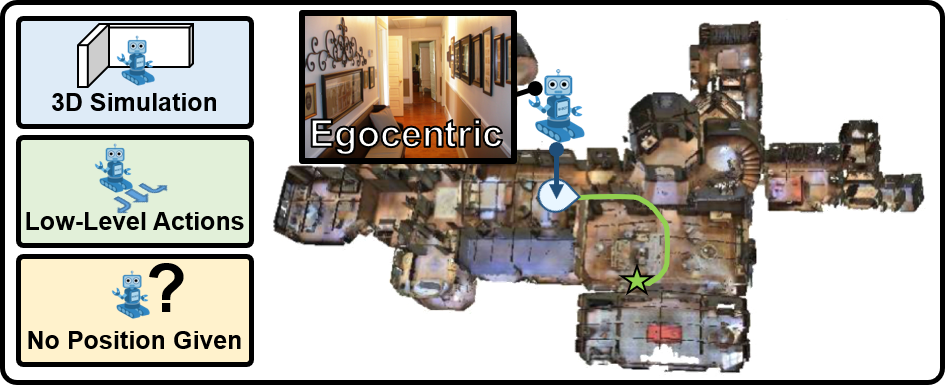}}
    \vspace{-1em}
    \caption{The VLN setting \textbf{(a)} operates on a fixed topology of panoramic images (shown in blue) -- assuming perfect navigation between nodes (often meters apart) and precise localization. Our VLN-CE setting \textbf{(b)} lifts these assumptions by instantiating the task in continuous environments with low-level actions -- providing a more realistic testbed for robot instruction following.}
    \label{fig:vlncmp}
\end{figure}

Many of these tasks have been developed from datasets of panoramic images captured in real scenes -- e.g. Google StreetView images in Touchdown \cite{chen2019touchdown} or Matterport3D panoramas captured in homes in Vision-and-Language Navigation (VLN) \cite{anderson2018vision}. This paradigm enables efficient data collection and high visual fidelity compared to 3D scanning or creating synthetic environments; however, scenes are only observed from a sparse set of points relative to the full 3D environment ($\sim$117 viewpoints per environment in VLN). As a consequence, environments in these tasks are defined in terms of a navigation graph (or \navg for short) -- a static topological representation of 3D space. As shown in \figref{fig:vln}, nodes in the \navg correspond to 360$^\degree$ panoramic images taken at fixed locations and edges between nodes indicate navigability. This \navg based formulation introduces a number of assumptions that make it a poor proxy for what a robotic agent would encounter while navigating the real world.  

Focusing our discussion on Vision-and-Language Navigation (VLN), the existence and common usage of the \navg imply the following assumptions:
\begin{compactitem}[--]
\item \textbf{Known topology.}
    Rather than continuous environments in which agents can move freely, agents operate on a fixed topology of traversable nodes (shown in blue in \figref{fig:vln}). Aside from being a poor match to robot control, this also provides prior information about environment layout to agents -- even in ``unseen'' test settings. For example, it is common practice to define agent actions by selecting directions in the current panorama and  `snapping' to the nearest adjacent \navg node in that direction. How an actual agent might acquire and update such a topology in new environments is an open question.
\item \textbf{Oracle navigation.}
    Movement between adjacent nodes in the nav-graph is deterministic, implying the existence of an oracle navigator capable of accurately traversing multiple meters in the presence of obstacles --  abstracting away the problem of visual navigation. Further, this movement between nodes is perceptually akin to teleportation -- the current panorama is simply replaced by the panorama at the new location meters away. This is in contrast to the continuous stream of observations a real agent would encounter while moving. 
\item \textbf{Perfect localization.}
    Agents are given their precise location and heading at all times. Most works use this data to encode precise geometry between nodes in the \navg as part of the decision making process, \eg moving ~30$^\degree$W and 1.12m forward from the previous node. Others use precise agent localization to construct spatial maps of the environment on which to reason about paths \cite{anderson2019chasing}. However, precise localization indoors is still a challenging problem.
\end{compactitem}
\noindent Taken together, these assumptions make current settings poor reflections of the real world both in terms of control (ignoring actuation, navigation, and localization error) and visual stimuli (lacking the poor framing and long observation-sequences agents will encounter). In essence, the problem is reduced to that of visually-guided graph search. As such, closing the loop by transferring these trained agents to physical robotic platforms has not been examined.

These assumptions are often justified by invoking existing technologies as potential oracles. For example, simultaneous localization and mapping (SLAM) or odometry systems can offer strong localization in appropriate conditions \cite{kohlbrecher11hector,mur2015orb}. Likewise, algorithms for path planning and control can navigate short distances in the presence of obstacles \cite{stentz1997optimal,gupta2017cognitive,wijmans2019dd}. Further, it is reasonable to suggest that issuing commands at the level of relative waypoints (in analogy to \navg nodes) is the proper interface between language-guided AI navigators and lower-level agent control. However, these techniques are each independently far from perfect and such an agent would need to learn the limitations of these lower-level control systems -- facing consequences when proposed waypoints cannot be reached effectively. Integrative studies such as these that combine and evaluate techniques for control and mapping with learned AI agents are not possible in current \navg based problem settings. 
In this work, we develop a continuous setting that enables these types of studies and take a first step towards integrating VLN agents with control via low-level actions.

\xhdr{Vision-and-Language Navigation in Continuous Environments.} In this work, we focus in on the Vision-and-Language Navigation (VLN) \cite{anderson2018vision} task and lift these implicit assumptions by instantiating it in continuous 3D environments rendered in a high-throughput simulator \cite{habitat19arxiv}. Consequently, we call this task Vision-and-Language Navigation in Continuous Environments (VLN-CE). Agents in our task are free to navigate to any unobstructed point through a set of low-level actions (e.g. move \texttt{forward} 0.25m, \texttt{turn-left} 15 degrees) rather than teleporting between fixed nodes. This setting introduces many challenges ignored in prior work. Agents in VLN-CE face significantly longer time horizons; the average number of actions along a path in VLN-CE is $\sim$55 compared to the 4-6 node hops in VLN (as illustrated in \figref{fig:vlncmp}). Moreover, the views the agent receives along the way are not well-posed by careful human operators as in the panoramas, but rather a consequence of the agent's actions. Agents must also learn to avoid getting stuck on obstacles, something that is structurally impossible in VLN's navigability defined \navg. Further, agents are not provided their location or heading while navigating.

We develop agent architectures for this task and explore how popular mechanisms for VLN transfer to the VLN-CE setting. Specifically, we develop a simple sequence-to-sequence baseline architecture as well as a cross-modal attention-based model.
We perform a number of input-modality ablations to assess the biases and baselines in this new setting (including models without perception or instructions as suggested in \cite{thomason2019shifting}). Unlike in VLN where depth is rarely used, our analysis reveals depth to be an integral signal for learning embodied navigation -- echoing similar findings in point-goal navigation tasks \cite{wijmans2019dd, habitat19arxiv}.  We also apply existing training augmentations \cite{ma2019self,ross2011reduction,tan2019learning} popular in VLN to our setting, finding mixed results. Overall, our best performing agent successfully navigates to the goal in approximately a third of episodes in unseen environments -- taking an average of 88 actions in this long-horizon task.

To further examine the relationship between the \navg-based VLN task and VLN-CE, we also transfer paths from agents trained in continuous environments back to the \navg to provide a direct comparison. We find significant gaps in performance between these settings indicative of the strong prior provided by the \navg. This suggests prior results in VLN may be overly optimistic in terms of progress towards instruction-following robots functioning in the wild.

\xhdr{Contributions.} To summarize our contributions, we:
\begin{compactitem}[\hspace{5pt}--]
\item Lift the VLN task to continuous 3D environments -- removing many unrealistic assumptions imposed by the \navg-based representation. The VLN-CE codebase and our baseline models are available at \url{https://github.com/jacobkrantz/VLN-CE}.\\[-0.8em]
\item Develop model architectures for the VLN-CE task and evaluate a suite of single-input ablations to assess the biases and baselines of the setting. \\[-0.8em]
\item Investigate how a number of popular techniques in VLN transfer to this more challenging long-horizon setting -- identifying significant gaps in performance.
\end{compactitem}

\csection{Related Work}
\label{sec:rel}

\noindent\textbf{Language-guided Visual Navigation Tasks.} Language-guided visual navigation tasks require agents to follow navigation directions in simulated environments. There have been a number of recent tasks proposed in this space \cite{anderson2018vision, chen2019touchdown,misra2018mapping,hermann2019learning}. Chen \etal\cite{chen2019touchdown} introduce the Touchdown task which studies outdoor language-guided navigation in Google Street View panoramas. Hermann \etal \cite{hermann2019learning} investigates the same setting; however, the instructions are automatically generated from Google Map directions rather than being crowdsourced from human annotators. Both adopt a \navg setting due to the source data being panoramic images -- constraining agent navigation to fixed points. Misra \etal \cite{misra2018mapping} introduce a simulated environment with unconstrained navigation and a dataset of crowdsourced instructions; however, the environments are unrealistic, synthetic scenes. Most related to our work is the Vision-and-Language Navigation (VLN) task of Anderson \etal\cite{anderson2018vision}. VLN provides \navg trajectories and crowdsourced instructions in Matterport3D \cite{Matterport3D} environments as the Room-to-Room (R2R) dataset. We build VLN-CE directly on these annotations -- converting R2R panorama-based trajectories to fine-grained paths in continuous Matterport3D environments (\figref{fig:vln} to \figref{fig:vlnotr}). As outlined in the introduction, this shift to continuous environments with unconstrained agent navigation lifts  a number of unrealistic assumptions.

\begin{table}[t]
    \centering
    \caption{Comparison of language-guided visual navigation tasks. Ours is the only to provide unconstrained navigation in real environments for crowdsourced instructions.}
    \resizebox{0.9\columnwidth}{!}{
    \setlength{\tabcolsep}{25pt}
    \begin{tabular}{l c c c}
        \toprule
        Task & Instructions & Environment & Navigation\\
        \midrule
         LANI \cite{misra2018mapping} &  Crowdsourced & Synthetic & Unconstrained\\
         StreetNav \cite{hermann2019learning} & Templated & Real & Nav-Graph Based \\
         Touchdown \cite{chen2019touchdown} & Crowdsourced & Real & Nav-Graph Based \\
         VLN \cite{anderson2018vision} & Crowdsourced & Real & Nav-Graph Based \\
         \midrule
         VLN-CE (ours) & Crowdsourced & Real & Unconstrained \\
         \bottomrule
    \end{tabular}}
    \label{tab:comparison}
\end{table}

The variation in these tasks is primarily in the source of navigation instructions (crowdsourced from human annotators vs.~generated via template), environment realism (hand-designed synthetic worlds vs.~captures from real locations), and constraints on agent navigation (\navg based navigation vs.~unconstrained agent motion). \tabref{tab:comparison} provides a comparison between tasks along these axes. Our proposed VLN-CE task provides the first setting with crowdsourced instructions in realistic environments with unconstrained agent navigation.

\xhdr{Approaches to Vision-and-Language Navigation.} VLN has seen considerable progress from a wide variety of techniques. Multimodal attention mechanisms have become popular to provide better grounding between instructions and the visual scene observation \cite{wang2019reinforced}. Orthogonal to new modeling architectures, improvements have also come from new training approaches and data augmentation methods. One prevalent technique is to utilize inverse ``speaker'' models to re-rank candidate trajectories or augment the available training data by generating instructions for novel trajectories~\cite{fried2018speaker}. Tan \etal~\cite{tan2019learning} further improve upon this idea by masking a subset of visual features during the speaker's instruction generation process, thereby improving the diversity of the generated instructions. Ma \etal\cite{ma2019self} show that an additional training signal can be gained by explicitly estimating progress toward the goal (referred to as self-monitoring). We adapt these methods to VLN-CE and examine their impact -- finding mixed results. Multimodal attention remains a useful structure; however, speaker-based data augmentation and self-monitoring losses provide mixed results.

\xhdr{Other Language-based Embodied AI.} A number of other embodied tasks have considered language-conditioned navigation. For instance, referring to specific rooms or objects that agents must then navigate to \cite{embodiedqa,gordon2018iqa,eqamatterport_cvpr_2019}. However, these settings use language to specify end-goals or query agent knowledge rather than to provide navigational directions. For example, specifying \myquote{lamp} or \myquote{What color is the lamp in the living room?} rather than \myquote{Go down the hall and into the bedroom on the right. Stop by the lamp to the left of the bed.} This loose coupling of intermediate agent action with the language instruction differentiates these tasks from language-guided navigation settings.

\csection{VLN in Continuous Environments (VLN-CE)}
\label{sec:data}

We consider a continuous setting for the vision-and-language navigation task which we refer to as Vision-and-Language Navigation in Continuous Environments (VLN-CE). Given a natural language navigation instruction, an agent must navigate from a start position to the described goal in a continuous 3D environment by executing a sequence of low-level actions based on egocentric perception alone. In overview, we develop this setting by transferring \navg-based Room-to-Room (R2R) \cite{anderson2018vision} trajectories to reconstructed continuous Matterport3D environments in the Habitat simulator \cite{habitat19arxiv}. We discuss the task specification and the details of this transfer process in this section.

\xhdr{Continuous Matterport3D Environments in Habitat.} We set our problem in the Matterport3D (MP3D) \cite{Matterport3D} dataset, a collection of 90 environments captured through over 10,800 high-definition RGB-D panoramas. In addition to the panoramic images, MP3D also provides corresponding mesh-based 3D environment reconstructions. To enable agent interaction with these meshes, we develop the VLN-CE task on top of the Habitat Simulator \cite{habitat19arxiv}, a high-throughput simulator that supports basic movement and collision checking for 3D environments including MP3D. In contrast to the simulator used in VLN \cite{anderson2018vision}, Habitat allows agents to navigate freely in the continuous environments.

\xhdr{Observations and Actions.} We select observation and action spaces to emulate a ground-based, zero-turning radius robot with a single, forward-mounted RGBD camera, similar to a LoCoBot \cite{locobot}. Agents perceive the world through egocentric RGBD images from the simulator with a resolution of 256$\times$256 and a horizontal field-of-view of 90 degrees. Note that this is similar to the egocentric RGB perception in the original VLN task \cite{anderson2018vision} but differs from the panoramic observation space adopted by nearly all follow-up work \cite{fried2018speaker,tan2019learning,ma2019self,wang2019reinforced}.

While the simulator is quite flexible in terms of agent actions, we consider four simple, low-level actions for agents in VLN-CE -- move \texttt{forward} 0.25m, \texttt{turn-left} or \texttt{turn-right} 15 degrees, or \texttt{stop} to declare that the goal position has been reached. These actions can easily be implemented on robotic agents with standard motion controllers. In contrast, actions to move between panoramas in \cite{anderson2018vision} traverse 2.25m on average and can include avoiding obstacles.

\csubsection{Transferring Nav-Graph Trajectories}

Rather than collecting a new dataset of trajectories and instructions, we instead transfer those from the nav-graph-based Room-to-Room dataset to our continuous setting. Doing so enables us to compare existing nav-graph-based techniques with our methods that operate in continuous environments on the same instructions. 

\xhdr{Matterport3D Simulator and the Room-to-Room Dataset.} The original VLN task is based on panoramas from Matterport3D (MP3D) \cite{Matterport3D}. To enable agent interaction with these panoramas, Anderson \etal \cite{anderson2018vision} developed the Matterport3D Simulator. Environments in this simulator are defined as nav-graphs $E = \{\mathcal{V},\mathcal{E}\}$. Each node $v \in \mathcal{V}$ corresponds to a panoramic image $I$ captured by a Matterport camera at location $x, y, z$ -- \ie $v = \{I, x, y, z\}$. Edges in the graph correspond to navigability between nodes. Navigability was defined by ray-tracing between node locations at varying heights to check for obstacles in the reconstructed MP3D scene and then manually inspected. Edges were manually added or removed based on judgement whether an agent could navigate between nodes -- including by avoiding minor obstacles\footnote{Details included from correspondence with the author of \cite{anderson2018vision}}. Agents act by teleporting between adjacent nodes in this graph. Based on this simulator, Anderson \etal \cite{anderson2018vision} collect the Room-to-Room (R2R) dataset containing 7189 trajectories each with three human-generated instructions on average. These trajectories consist of a sequence of nodes $\tau=[v_1,\dots,v_T]$ with length $T$ averaging between 4 and 6 nodes.

\begin{figure}[t]
    \centering
    \includegraphics[width=\textwidth]{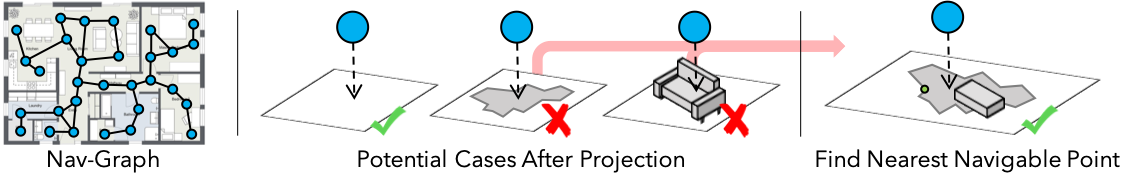}
    \caption{We transfer nav-graph trajectories over panoramas (blue dots) from the Room-to-Room (R2R) dataset to locations in reconstructed Matterport3D (MP3D) environments. Some map to `holes' in environment meshes where reconstruction failed or on furniture (commonly tables) where an agent could not navigate. For these, we find the nearest navigable point within 0.5m.}
    \label{fig:projection}
\end{figure}

\xhdr{Converting Room-to-Room Trajectories to Habitat.} Given a mapping between the coordinate frames of Matterport3D Simulator and MP3D in Habitat, it is seemingly simple to transfer the Room-to-Room trajectories -- after all, each node has a corresponding $xyz$ location. However, node locations often do not correspond to reachable locations for a ground-based agent -- existing at variable height depending on tripod configuration or placed on top of flat furniture like tables. Further, the reconstructions and panoramas may differ if objects or doors are moved between camera captures. \figref{fig:projection} shows an overview of this process and common errors when directly transferring node locations.

\begin{figure}[t]
    \centering
    \subfigure[\scriptsize Node Location Displacement\label{fig:node_error}]{
        \includegraphics[clip=true, trim=15pt 15pt 12pt 12pt, width=0.31\textwidth]{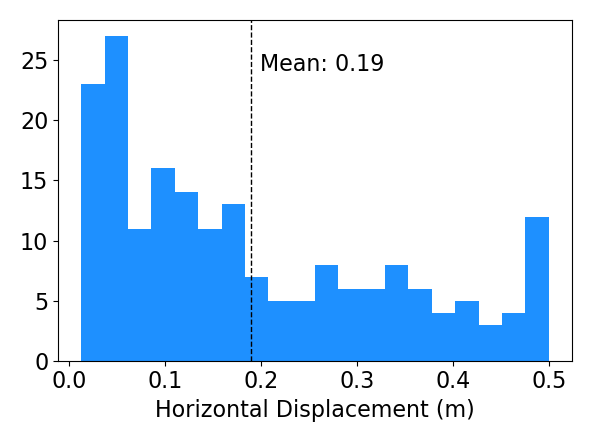}
    }\hfill%
    \subfigure[\scriptsize Discontinuities
    \label{fig:errors}]{
        \includegraphics[width=0.27\textwidth]{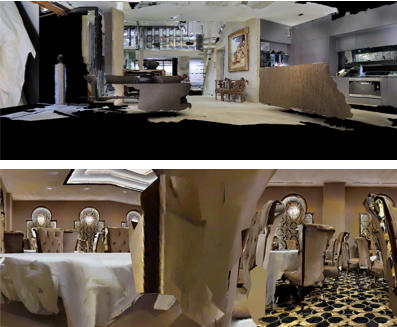}
    }%
    \hfill%
    \subfigure[\scriptsize Trajectory Length in Actions\label{fig:length}]{
        \includegraphics[clip=true, trim=15pt 15pt 12pt 12pt, width=0.31\textwidth]{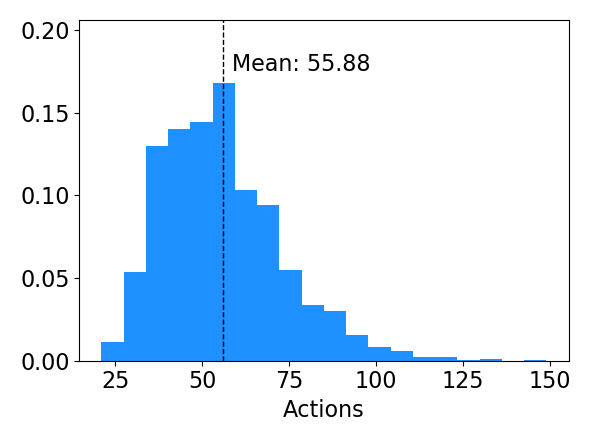}
    }
    \caption{We successfully transfer 77\% of the R2R trajectories. (a) Most panorama nodes transfer directly, but 3\% require horizontal adjustment -- with an average displacement of 0.19m. (b) Despite this, some trajectories are not navigable because of differences between the panoramas and reconstructed environments, \eg holes in the reconstructed mesh (top) or objects like chairs being manipulated between panorama captures (bottom).  (c) Our setting requires significantly more agent decisions per trajectory with an average action length of 55.88 compared to 5 in R2R.}
    \label{fig:my_label}
\end{figure}

For each node, $v=\{I, x, y, z\}$, we would like to identify the nearest, navigable point on the reconstructed mesh -- \ie the closest point that can be occupied by a ground-based agent represented by a 1.5m tall cylinder of diameter of 0.2m. Directly projecting to the nearest mesh location fails for 73\% of nodes where failure is defined as projecting to distant (>0.5m) or non-navigable points. Many of these points project to ceilings or the tops of nearby objects rather than the floor due to the height of the camera. Instead, we cast a ray up to 2m directly downward from the node location. At small, fixed intervals along this ray, we project to the nearest mesh point. If multiple navigable points are identified, we take the one with minimal horizontal displacement from the original location. If no navigable point is found with less than a 0.5m displacement, we consider this MP3D node unmappable to the 3D mesh and thus invalid. We reviewed all invalid nodes manually and made corrections if possible, \eg shifting nodes to the side of furniture. After these steps, 98.3\% of nodes are successfully transferred. We refer to these transferred nodes as waypoint locations in the MP3D environments. As shown in \figref{fig:node_error}, points requiring adjustment (3\% of points) are transferred with small horizontal displacement, averaging 0.19m from the panorama location.

Given a trajectory of converted waypoints $\tau=[w_1,\dots,w_T]$, we would like to verify that an agent can actually navigate between each location. We employ an A*-based heuristic search algorithm to compute an approximate shortest path to a goal location. We run this shortest path algorithm between each waypoint in a trajectory to the next (\eg $w_i$ to $w_{i+1}$). A trajectory is considered navigable if for each pairwise navigation, an agent following the computed shortest path can navigate to within 0.5m of the next waypoint ($w_{i+1}$). In total, we find 77\% of the R2R trajectories navigable in the continuous environment.

\xhdr{Non-Navigable Trajectories.} Among the 23\% of trajectories that were not navigable, we observed two primary failure modes. First and most simply, {22\%} of these included one of the 1.7\% of invalid nodes that could not be projected to MP3D 3D meshes and were rejected by default. The remaining trajectories were not navigable because they spanned disjoint regions of the reconstruction -- meaning that there was no valid path from some waypoint $w_i$ to $w_{i+1}$. As shown in \figref{fig:errors}, this may be holes or other mesh errors dividing the space. Alternatively, objects like chairs may be moved in between panorama captures -- possibly resulting in a reconstruction that places the object mesh on top of individual panorama locations. As noted above, nodes in the R2R nav-graph were manually connected if there appeared to be a path between them, even if most other panoramas (and thus the reconstruction) showed objects (\eg a closed door) blocking their path.

\csubsection{VLN-CE Dataset}
In total, the VLN-CE dataset consists of 4475 trajectories converted from R2R train and validation splits. For each trajectory, we provide the multiple natural language instructions from R2R and a pre-computed shortest path following the waypoints via low-level actions. As shown in \figref{fig:length}, the low-level action space of VLN-CE makes our trajectories significantly longer horizon tasks -- with an average of 55.88 steps compared to the 4-6 in R2R.

\csection{Instruction-guided Navigation Models in VLN-CE}
\label{sec:approach}

We develop two models for VLN-CE. A simple sequence-to-sequence baseline and a more powerful cross-modal attentional model. While there are many differences in the details, these models are conceptually similar to early \cite{anderson2018vision} and more recent \cite{wang2019reinforced} work in the \navg based VLN task. Exploring these gives insight into the difficulty of this setting in isolation and by comparison relative to VLN. Further, these models allow us to test whether improvements from early to later architectures carry over to a more realistic setting. Both of our models make use of the same observation and instruction encodings described below.

\xhdr{Instruction Representation.} We convert tokenized instructions to corresponding GLoVE \cite{pennington_emnlp14} embeddings which are processed by recurrent encoders for each model. We denote these encoded tokens as $\mathbf{w}_1, \dots, \mathbf{w}_T$ for a length $T$ instruction.

\xhdr{Observation Encoding.} We separately encode the RGB and depth observations. For RGB, we apply a ResNet50\cite{he_cvpr16} pretrained on ImageNet \cite{imagenet_cvpr09} to collect semantic visual features. We denote the final spatial features of this model as $\mathcal{V} = \{\mathbf{v}_{i}\}$ where $i$ indexes over spatial locations. Likewise for depth, we use a modified ResNet50 that was trained to perform point-goal navigation (\ie to navigate to a location given in relative coordinates) \cite{wijmans2019dd} and denote these as $\mathcal{D} = \{\mathbf{d}_{i}\}$.

\begin{figure}[t]
    \centering
    \includegraphics[width=0.031\textwidth]{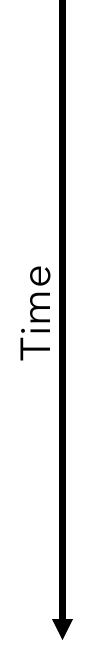}\hfill%
    \subfigure[\scriptsize Sequence-to-Sequence Baseline  \label{fig:base}]{
    \includegraphics[clip=true, trim=0pt 6pt 6pt 7pt, height=1.2in, width=0.37\textwidth]{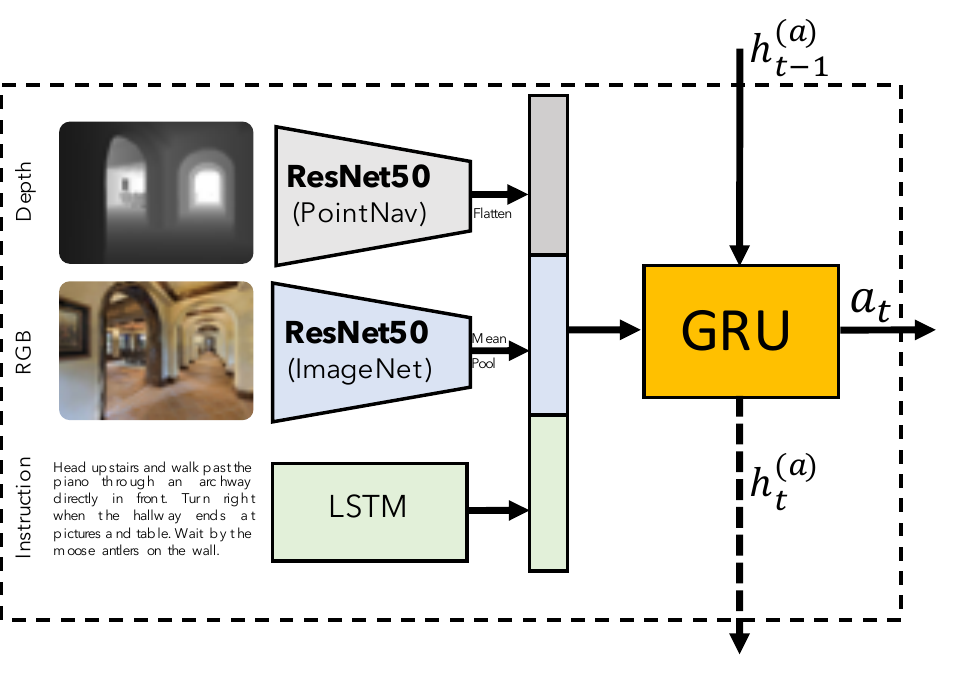}}\hfill
    \subfigure[\scriptsize Cross-Modal Attention Model \label{fig:attn}]{
        \includegraphics[clip=true, trim=0pt 6pt 6pt 6pt, height=1.2in,width=0.58\textwidth]{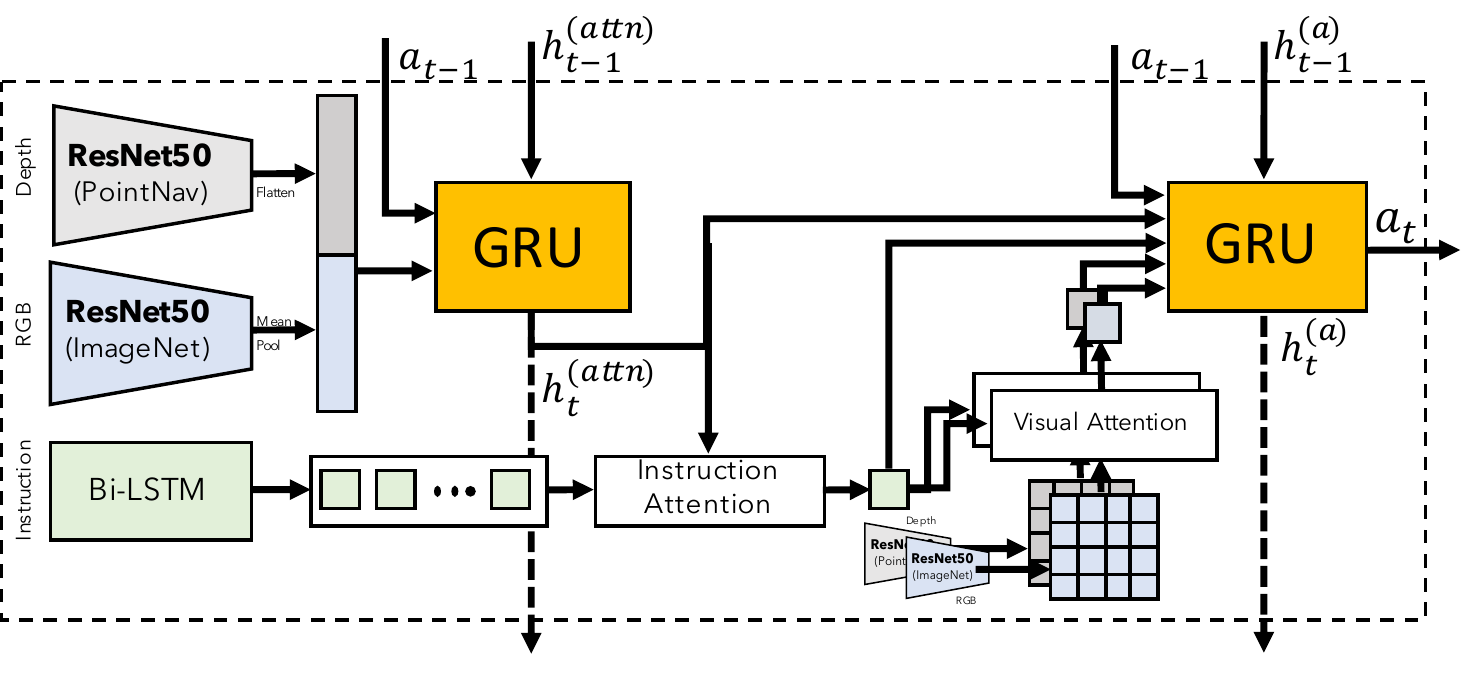}}
    \caption{We develop a simple baseline agent (a) as well as an attentional agent (b) comparable to that in \cite{wang2019reinforced}. Both receive RGB and depth frames represented by pretrained networks for image classification \cite{imagenet_cvpr09} and point-goal navigation \cite{wijmans2019dd}, respectively.} 
    \label{fig:models}
\end{figure}

\csubsection{Sequence-to-Sequence Baseline}
We consider a simple sequence-to-sequence baseline model shown in \figref{fig:base}. This model consists of a recurrent policy that takes a representation of the visual observation (depth and RGB) and instructions at each time step, then predicts an action $a$. Concretely, we can write the agent for time step $t$ as 
\begin{equation}
\bar{\mathbf{v}}_t = \mbox{mean-pool}\left(\mathcal{V}_t\right), ~~ \bar{\mathbf{d}}_t = \left[\mathbf{d}_{1}, \dots, \mathbf{d}_{wh}\right], ~~ \mathbf{s} = \mbox{LSTM}\left(\mathbf{w}_1, \dots, \mathbf{w}_T\right)
\end{equation}
\begin{equation}
\mathbf{h}_t^{(a)} = \mbox{GRU}\left( \left[\bar{\mathbf{v}}_t,\bar{\mathbf{d}}_t,\mathbf{s}\right], \mathbf{h}^{(a)}_{t-1}\right)
\end{equation}
\begin{equation}
a_t = \argmax_a~~ \mbox{softmax}\left(W_a\mathbf{h}^{(a)}_t + \mathbf{b}_a\right)~
\end{equation}
where $[\cdot]$ denotes concatenation and $\mathbf{s}$ is the final hidden state of an LSTM instruction encoder. This simple model enables straight-forward input-modality ablations and establishes a straight-forward baseline for the VLN-CE setting.

\csubsection{Cross-Modal Attention Model}
While the previous baseline is a sensible start, it lacks powerful modeling techniques common to vision-and-language tasks including cross-modal attention and spatial visual reasoning which are intuitively quite important for language-guided visual navigation. Many instructions include relative references (\eg \myquote{to the left of the table}) that would be difficult to ground from mean-pooled features. Moreover, already-completed parts of the instruction are likely irrelevant to the next decision -- pointing towards the potential of attention over instructions.

We consider a more expressive model shown in \figref{fig:attn} that incorporates these mechanisms. This model consists of two recurrent networks -- one tracking visual observations as before and the other making decisions based on attended instruction and visual features. We can write this first recurrent network as:

\begin{equation}
\mathbf{h}_t^{(attn)} = \mbox{GRU}\left( \left[\bar{\mathbf{v}}_t,\bar{\mathbf{d}}_t,\mathbf{a}_{t-1}\right], \mathbf{h}^{(attn)}_{t-1}\right)
\end{equation}

\noindent where $\mathbf{a}_{t-1} \in \mathbb{R}^{1 \times 32}$ and is a learned linear embedding of the previous action.
We encode instructions with a bi-directional LSTM and reserve all intermediate hidden states: 
\begin{equation}
\mathcal{S}=\{\mathbf{s_1},\dots, \mathbf{s_T}\} = \mbox{BiLSTM}\left(\mathbf{w}_1, \dots, \mathbf{w}_T\right)
\end{equation}
\noindent We then compute an attended instruction feature $\hat{\mathbf{s}}_t$ over these representations which is then used to attend to visual ($\hat{\mathbf{v}}_t$) and depth ($\hat{\mathbf{d}}_t$) features. Concretely, 
\begin{equation}
    \hat{\mathbf{s}}_t = \mbox{Attn}\left(\mathcal{S}, \mathbf{h}_t^{(attn)}\right),~~ \hat{\mathbf{v}}_t = \mbox{Attn}\left(\mathcal{V}_t, \hat{\mathbf{s}}_t\right), ~~ \hat{\mathbf{d}}_t = \mbox{Attn}\left(\mathcal{D}_t, \hat{\mathbf{s}}_t\right)
\end{equation}

\noindent where Attn is a scaled dot-product attention \cite{vaswani2017attention}. For a query $\mathbf{q} \in \mathbb{R}^{1\times d_q}$, $\hat{\mathbf{x}}=\mbox{Attn}(\{\mathbf{x}_i\}, \mathbf{q})$ is computed as $\hat{\mathbf{x}}{=} \sum_i \alpha_i \mathbf{x}_i$ for  $\alpha_i{=}\mbox{softmax}_i((W_K \mathbf{x}_i)^T \mathbf{q}~/ \sqrt{d_q})$. The second recurrent network then takes a concatenation of these features as input (including an action encoding and the first recurrent network's hidden state) and predicts an action.

\begin{equation}
\mathbf{h}_t^{(a)} = \mbox{GRU}\left( \left[\hat{\mathbf{s}}_t, \hat{\mathbf{v}}_t, \hat{\mathbf{d}}_t, \mathbf{a}_{t-1}, \mathbf{h}^{(attn)}_t\right], \mathbf{h}^{(a)}_{t-1}\right)
\end{equation}

\begin{equation}
a_t = \argmax_a~~ \mbox{softmax}\left(W_a\mathbf{h}^{(a)}_t + \mathbf{b}_a\right)~
\end{equation}

\csubsection{Auxiliary Losses and Training Regimes}

Aside from modeling details, much of the remaining progress in VLN has come from adjusting the training regime -- adding auxiliary losses / rewards \cite{ma2019self, wang2019reinforced}, mitigating exposure bias during training \cite{anderson2018vision,wang2019reinforced}, or reducing data sparsity by incorporating synthetically generated data augmentation \cite{tan2019learning,fried2018speaker}. We explore some of these directions for VLN-CE, but note that this is not an exhaustive accounting of impactful techniques. Particularly, we suspect that methods addressing exposure bias and data sparsity in VLN will help in the VLN-CE setting where these problems may be amplified by lengthy action sequences. We report ablations with and without these techniques in \secref{sec:exp}.

\xhdr{Imitation Learning.} A natural starting point for training is simply to maximize the likelihood of the ground truth trajectories. To do so, we perform teacher-forcing training with inflection weighting. As described in \cite{eqamatterport_cvpr_2019}, inflection weighting places emphasis on time-steps where actions change (\ie $a_{t-1} \neq a_t$), adjusting loss weight proportionally to the rarity of such events.  This was found to be helpful for problems like navigation with long sequences of repeated actions (\eg going forward down a hall). We observe a similar effect in early experiments and apply inflection weighting in all our experiments. 

\xhdr{Coping with Exposure Bias.} Imitation learning in auto-regressive settings suffers from a disconnect between training and test -- agents are not exposed to the consequences of their actions during training. Prior work has shown significant gains by addressing this issue for VLN through scheduled sampling \cite{anderson2018vision} or reinforcement learning fine-tuning \cite{wang2019reinforced, tan2019learning}. In this work, we apply Dataset Aggregation (DAgger) \cite{ross2011reduction} towards the same end. While DAgger and scheduled sampling share many similarities, DAgger trains on the aggregated set of trajectories from all iterations 1 to $n$. Thus, the resulting policy after iteration $n$ is optimized over all past experiences and not just those collected from iteration $n$.

\xhdr{Synthetic Data Augmentation.} Another popular strategy is to learn an inverse `speaker' model that produces instructions given a trajectory. These models can be used to re-rank paths or to generate new trajectory-instruction pairs from any trajectory. Both \cite{tan2019learning} and \cite{fried2018speaker} take this data augmentation approach and many follow-up works have used these trajectories for gains in performance. We take the $\sim$150k synthetic trajectories generated this way from \cite{tan2019learning} -- converting them to our continuous environments. 

\xhdr{Progress Monitor.} An important aspect of a successful navigation is accurately identifying where to stop. Prior work \cite{ma2019self} has found improvements from explicitly supervising the agent with a progress-toward-goal signal. Specifically, agents are trained to predict the fraction through the trajectory they are at each time step. We apply this progress estimation during training with a mean squared error loss term akin to \cite{ma2019self}.

\csection{Experiments}
\label{sec:exp}

\noindent\textbf{Setting and Metrics.} We train and evaluate our models in VLN-CE. As is common practice, we perform early stopping based on val-unseen performance. We report standard metrics for visual navigation tasks defined in \cite{anderson2018vision,anderson2018evaluation,magalhaes2019effective} -- trajectory length in meters (\texttt{TL}), navigation error in meters from goal at termination (\texttt{NE}), oracle success rate (\texttt{OS}), success rate (\texttt{SR}), success weighted by inverse path length (\texttt{SPL}), and normalized dynamic-time warping (\texttt{nDTW}). For our discussion, we will examine success rate and SPL as the primary metrics for performance and use NDTW to describe how paths differ in shape from ground truth trajectories. For full details on these metrics, see \cite{anderson2018vision,anderson2018evaluation,magalhaes2019effective}.

\xhdr{Implementation Details.}  We utilize the Adam optimizer~\cite{kingma_iclr15} with a learning rate of $2.5\times10^{-4}$ and a batch size of 5 full trajectories.  We set the inflection weighting coefficient~\cite{eqamatterport_cvpr_2019} to $3.2$ (inverse frequency of inflections in our ground-truth paths). We train on all ground-truth paths until convergence on val-unseen (at most 30 epochs).  For DAgger~\cite{ross2011reduction}, we collect the $n$th set by taking the oracle action with probability $\beta{=}0.75^{n}$ and the current policy action otherwise.  We collect $5,000$ trajectories at each stage and then perform 4 epochs of imitation learning (with inflection weighting) over all collected trajectories.  Once again, we train to convergence on val-unseen (6 to 10 dataset collections, depending on the model).  We implement our agents in PyTorch~\cite{paszke2019pytorch} and on top of Habitat~\cite{habitat19arxiv}. 

\begin{table}[t]
	\caption{No-learning baselines and input modality ablations for our baseline sequence-to-sequence model. Given the long trajectories involved, we find both random agents and single-modality ablations to perform quite poorly in VLN-CE.}\vspace{-2.5em}
	\label{tab:ablate}
	\setlength{\tabcolsep}{3pt}
	\begin{center}
		\resizebox{\columnwidth}{!}{
		\begin{tabular}{l ccc c cccccs c cccccs}
			\midrule

			& & &  & & \multicolumn{6}{c}{\scriptsize\textbf{Val-Seen}} & & \multicolumn{6}{c}{\scriptsize\textbf{Val-Unseen}}  \\
			\cmidrule{6-11} \cmidrule{13-18}
			\scriptsize Model & \scriptsize Vision & \scriptsize Instr. & \scriptsize History & & \scriptsize\textbf{\texttt{TL}}~$\downarrow$ & \scriptsize\textbf{\texttt{NE}}~$\downarrow$ & \scriptsize\textbf{\texttt{nDTW}}~$\uparrow$ & \scriptsize\textbf{\texttt{OS}}~$\uparrow$ & \scriptsize\textbf{\texttt{SR}}~$\uparrow$ &\scriptsize\textbf{\texttt{SPL}}~$\uparrow$ & & \scriptsize\textbf{\texttt{TL}}~$\downarrow$ & \scriptsize\textbf{\texttt{NE}}~$\downarrow$ & \scriptsize\textbf{\texttt{nDTW}}~$\uparrow$ & \scriptsize\textbf{\texttt{OS}}~$\uparrow$ & \scriptsize\textbf{\texttt{SR}}~$\uparrow$ &\scriptsize\textbf{\texttt{SPL}}~$\uparrow$\\
			\midrule
			Random & - & - & - && 3.54 & 10.20 & 0.28 & 0.04 & 0.02 & 0.02&  & 3.74 & 9.51 & 0.30 & 0.04 & 0.03 & 0.02\\
			Hand-Crafted & - & - & - && 3.83 & 9.56 & 0.33 & 0.05 & 0.04 & 0.04&  & 3.71 & 10.34 & 0.30 & 0.04 & 0.03 & 0.02\\
			\midrule
            \basemodel & \scriptsize \texttt{RGBD} & \checkmark & \checkmark && 8.40 & 8.54 & 0.45 & 0.35 & 0.25 & 0.24 &  & 7.67 & 8.94 & 0.43 & 0.25 & 0.20 & 0.18 \\
            \hspace{10pt} -- No Image & \scriptsize \texttt{D} & \checkmark &  \checkmark && 7.77 & 8.55 & 0.46 & 0.31 & 0.24 & 0.23  && 7.87 & 9.09 & 0.41 & 0.23 & 0.17 & 0.15\\
            \hspace{10pt} -- No Depth & \scriptsize \texttt{RGB} & \checkmark &  \checkmark && 4.93 & 10.76 & 0.29 & 0.10 & 0.03 & 0.03 &  &5.54 & 9.89 & 0.31 & 0.11 & 0.04 & 0.04\\
            \hspace{10pt} -- No Vision & - & \checkmark & \checkmark && 4.26 & 11.07 & 0.26 & 0.03 & 0.00 & 0.00 &  &4.68 & 10.06 & 0.30 & 0.07 & 0.00 & 0.00\\
            \hspace{10pt} -- No Instruction & \scriptsize \texttt{RGBD} & - &  \checkmark && 7.86 & 9.09 & 0.42 & 0.26 & 0.18 & 0.17  && 7.27 & 9.03 & 0.42 & 0.22 & 0.17 & 0.16\\
            \bottomrule
		\end{tabular}}
	\end{center}
\end{table}

\csubsection{Establishing Baseline Performance for VLN-CE}

\xhdr{No-Learning Baselines.} To establish context for our results, we consider random and hand-crafted agents shown in \tabref{tab:ablate} (top two rows). The {\texttt{random}} agent selects actions according to the train set action distribution (68\% \texttt{forward}, 15\% \texttt{turn-left}, 15\% \texttt{turn-right}, and 2\% \texttt{stop}). The {\texttt{hand-crafted}} agent picks a random heading and takes 37 \texttt{forward} actions (average trajectory length) before calling \texttt{stop}. Despite having no learned components nor processing any input, both these agents achieve approximately 3\% success rates in val-unseen. In contrast, a similar hand-crafted random-heading-and-forward model in VLN yields a 16.3\% success rate \cite{anderson2018vision}. Though not directly comparable, this gap illustrates the strong structural prior provided by the nav-graph in VLN. 

\xhdr{Seq2Seq and Single-Modality Ablations.} \tabref{tab:ablate} also shows performance for the baseline Seq2Seq model along with input ablations. All models are trained with imitation learning without data augmentation or any auxiliary losses. Our baseline Seq2Seq model significantly outperforms the random and hand-crafted baselines, successfully reaching the goal in 20\% of val-unseen episodes. 

As illustrated in \cite{thomason2019shifting}, models examining only single modalities can be very strong baselines in embodied tasks. We train models without access to the instruction (\texttt{No Instruction}) and with ablated visual input (\texttt{No Vision/Depth/Image}). 
All of these ablations under-perform the Seq2Seq baseline. We find that depth is a very strong signal for learning, with models lacking it (\texttt{No Depth} and \texttt{No Vision}) failing to outperform chance ($\leq$1\% success rates). We believe that depth enable agents to quickly begin traversing environments effectively (\eg without collisions) and without this it is very difficult to bootstrap to instruction following. With a success rate of 17\%, the \texttt{No Instruction} model performs similarly to a hand-crafted agent in VLN, suggesting shared trajectory regularities between VLN and VLN-CE. While these regularities can be manually exploited in VLN via the nav-graph, they are implicit in VLN-CE as evidenced by the significantly lower performance of our random and hand crafted agents which collide with and get stuck on obstacles. The \texttt{No Image} model also achieves 17\% success, similarly failing to reason about instructions. This hints at the importance of grounding visual referents (through RGB) for navigation. 

\setlength{\tabcolsep}{.3em}
\begin{table}[t]
	\caption{Performance in VLN-CE. We find that popular techniques in VLN have mixed benefit in VLN-CE; however, our best performing model combining all examined techniques succeeds nearly 1/3rd of the time in new environments. {\scriptsize \textsuperscript{*} denotes fine-tuning.}
	}\vspace{-25pt}
	\label{tab:results}
	\begin{center}
	    \resizebox{\textwidth}{!}{
		\begin{tabular}{l l ccc c cccccs c cccccs}
			\midrule

			&& \multirow{2}{*}[-1em]{\scriptsize \shortstack{PM\\ \cite{ma2019self}}}&\multirow{2}{*}[-1em]{\scriptsize \shortstack{DA\\\cite{ross2011reduction}}} & \multirow{2}{*}[-1em]{\scriptsize  \shortstack{Aug.\\ \cite{tan2019learning}}} & & \multicolumn{6}{c}{\scriptsize\textbf{Val-Seen}} & & \multicolumn{6}{c}{\scriptsize\textbf{Val-Unseen}}  \\
			\cmidrule{7-12} \cmidrule{14-19}
			\scriptsize \texttt{\#} &\scriptsize Model &    &  && & \scriptsize\textbf{\texttt{TL}}~$\downarrow$ & \scriptsize\textbf{\texttt{NE}}~$\downarrow$ & \scriptsize\textbf{\texttt{nDTW}}~$\uparrow$ & \scriptsize\textbf{\texttt{OS}}~$\uparrow$ & \scriptsize\textbf{\texttt{SR}}~$\uparrow$ &\scriptsize\textbf{\texttt{SPL}}~$\uparrow$ & & \scriptsize\textbf{\texttt{TL}}~$\downarrow$ & \scriptsize\textbf{\texttt{NE}}~$\downarrow$ &
			\scriptsize\textbf{\texttt{nDTW}}~$\uparrow$ &
			\scriptsize\textbf{\texttt{OS}}~$\uparrow$ & \scriptsize\textbf{\texttt{SR}}~$\uparrow$ & \scriptsize\textbf{\texttt{SPL}}~$\uparrow$\\
			\midrule

			\scriptsize \texttt{1} & \multirow{5}{*}{\shortstack[l]{\basemodel\\Baseline}} & - & -& -&& 8.40 & 8.54 & 0.45 & 0.35 & 0.25 & 0.24  &  & 7.67 & 8.94 & 0.43& 0.25 & 0.20 & 0.18  \\
			 \scriptsize \texttt{2} & & \checkmark & - & - && 8.34 & 8.48 & 0.47 & 0.32 &0.22  & 0.21&  & 8.93 & 9.28 & 0.40 & 0.28 & 0.17 & 0.15 \\
			 \scriptsize \texttt{3} && - & \checkmark & - && 9.32 & 7.09 & 0.53 & 0.44 & 0.34 &0.32  &  & 8.46 & 7.92 & 0.48 & 0.35 & 0.26 & 0.23 \\
			\scriptsize \texttt{4} && - & - & \checkmark && 8.23 & 7.76 & 0.51 & 0.34 & 0.26 &0.25  &  & 7.22 & 8.70 & 0.44 & 0.26 & 0.19 & 0.17 \\
			\scriptsize \texttt{5} & & \checkmark & \checkmark\textsuperscript{*} & \checkmark && 9.37 & 7.02 & 0.54 & 0.46 & 0.33 & 0.31&  & 9.32 & 7.77 & 0.47 & 0.37 & 0.25 & 0.22 \\
			\midrule
			\scriptsize \texttt{6} &\multirow{7}{*}{ \shortstack[l]{Cross-Modal\\ Attention}} & - & -& -&& 8.26 & 7.81 & 0.49 & 0.38 & 0.27 & 0.25 &  & 7.71 & 8.14 & 0.47 & 0.31 & 0.23 & 0.22  \\
			\scriptsize \texttt{7} & & \checkmark & - & - && 8.51 & 8.17 & 0.47 & 0.35 & 0.28 &0.26   &  & 7.87 & 8.72 & 0.44 & 0.28 & 0.21 & 0.19 \\
			\scriptsize \texttt{8} & & - & \checkmark & - && 8.90 & 7.40 & 0.52 & 0.42 & 0.33 &0.31  &  & 8.12 & 8.00 & 0.48 & 0.33 & 0.27 & 0.25 \\
			\scriptsize \texttt{9} && - & - & \checkmark && 8.50 & 8.05 & 0.49 & 0.36 &0.26  & 0.24 &  & 7.58 & 8.65 & 0.45 & 0.28 & 0.21 & 0.19 \\
			\scriptsize \texttt{10} & & \checkmark & \checkmark\textsuperscript{*} & \checkmark && 9.26 & 7.12 & 0.54 & 0.46 & \textbf{0.37} & \textbf{0.35} &  & 8.64 & \textbf{7.37} & \textbf{0.51} & \textbf{0.40} & \textbf{0.32} & \textbf{0.30} \\
			\cmidrule{3-19}
			\scriptsize \texttt{11} && \checkmark & - & \checkmark && 8.49 & 8.29 & 0.47 & 0.36 &0.27  & 0.25 &  & 7.68 & 8.42 & 0.46 & 0.30 & 0.24 & 0.22 \\
			\scriptsize \texttt{12} && - & \checkmark\textsuperscript{*} & \checkmark && 9.32 & \textbf{6.76} & \textbf{0.55} & \textbf{0.47} &\textbf{0.37}  & 0.33 &  & 8.27 & 7.76 & 0.50 & 0.37 & 0.29 & 0.26 \\
			\bottomrule
		\end{tabular}}
	\end{center}
\end{table}

\csubsection{Model Performance in VLN-CE}
\tabref{tab:results} shows a comparison of our models (\texttt{Seq2Seq} and \texttt{Cross-Modal}) under three training augmentations (\texttt{Progress Monitor}, \texttt{DAgger}, \texttt{Data Augmentation}). 

\xhdr{Cross-Modal Attention \vs~Seq2Seq.} We find the cross-modal attention model outperforms the Seq2Seq baseline under all settings for new environments. For example, in teacher-forcing training (row \texttt{1} \vs~\texttt{6}), the cross-modal attention model improves from 0.18 to 0.22 SPL on val-unseen, an improvement of 0.04 SPL (22\% relative). Likewise, when applying all three augmentations (row \texttt{5} \vs~\texttt{10}), the cross-modal model improves from 0.22 to 0.30 SPL, an improvement of 0.08 SPL (36\% relative).

\xhdr{Training Augmentation.} We find DAgger-based training impactful for both the Seq2Seq (row \texttt{1} \vs~\texttt{3}) and Cross-Modal (row \texttt{6} \vs~\texttt{8}) models -- improving by 0.03-0.05 SPL in val-unseen. Contrary to findings in prior work, we observe negative effects from progress monitor auxiliary loss or data augmentation for both models (rows \texttt{2/4} and \texttt{7/9}) -- dropping 0.01-0.03 SPL from standard training (rows \texttt{1/6}). 
Despite this, we find combining all three techniques to lead to significant performance gains for the cross-modal attention model (row \texttt{10}). Specifically, we pretrain with imitation learning, data augmentation, and the progress monitoring loss, then finetune using DAgger (with $\beta{=}0.75^{n+1}$) on the original data. This Cross-Modal Attention PM+DA\textsuperscript{*}+Aug model achieves an SPL of 0.35 on val-seen and 0.30 on val-unseen -- succeeding on 32\% of episodes in new environments.

We explore this trend further for the Cross-Modal model. We examine the validation performance of PM+Aug (row \texttt{11}) and find it to outperform Aug or PM alone (by 0.02-0.03 SPL).  Next, we examine progress monitor loss on val-unseen for both PM and PM+Aug.  We find that without data augmentation, the progress monitor over-fits considerably more (validation loss of 0.67 \vs 0.47) -- indicating that the progress monitor can be effective in our continuous setting but tends to over-fit on the non-augmented training data, negatively affecting generalization.  Finally, we examine the performance of DA\textsuperscript{*}+Aug (row \texttt{12}) and find that this outperforms DA (by 0.01-0.02 SPL), but is unable to match pre-training with the progress monitor and augmented data (row \texttt{10}).

\begin{figure}[t]
    \centering
    \includegraphics[width=0.9\textwidth]{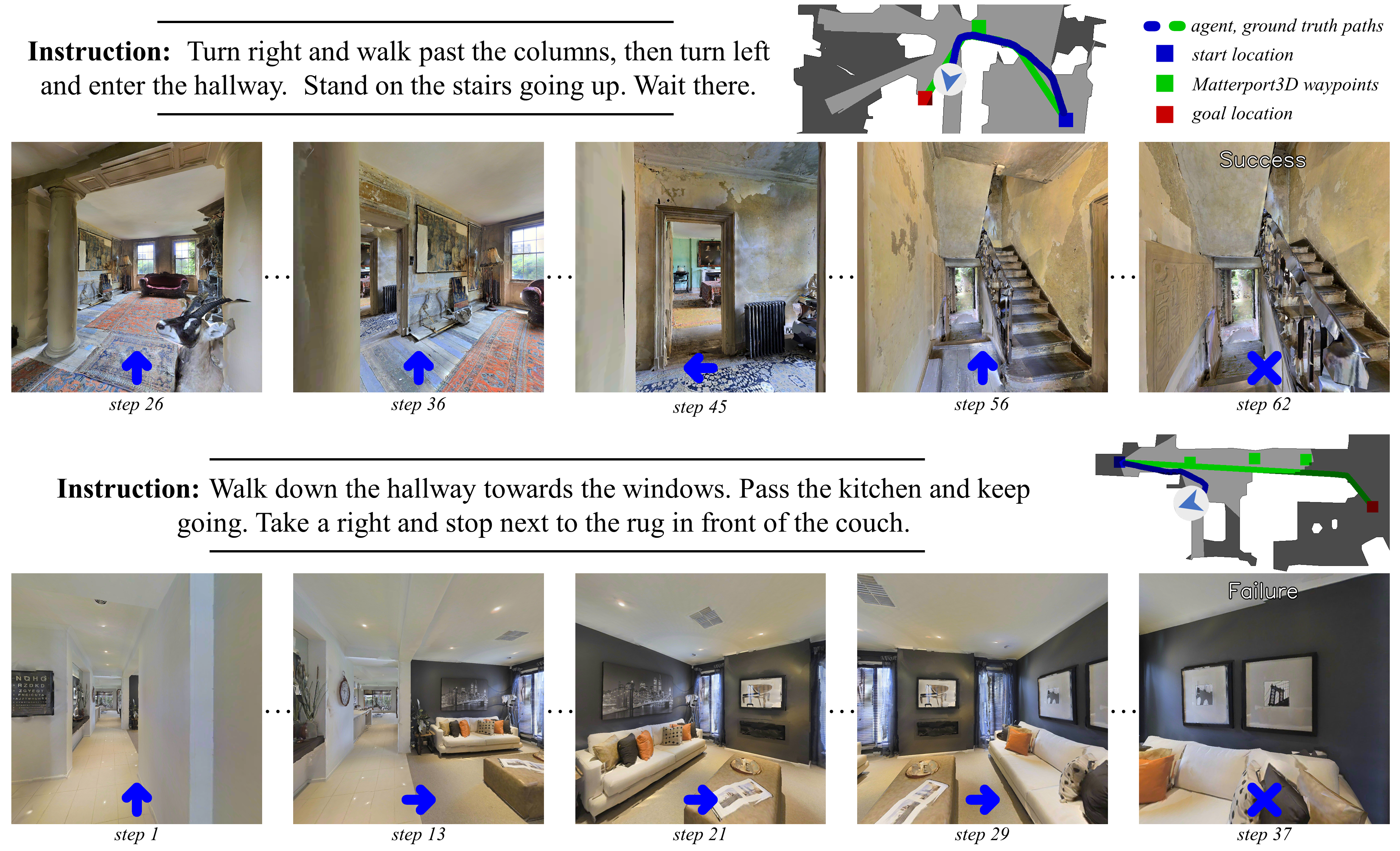}
    \caption{Qualitative examples of our Cross Modal Attention model taken in unseen validation environments. In the first example our agent successfully follows the instruction -- note it takes 62 actions in VLN-CE, whereas the VLN traversal requires just 3 hops.}
    \label{fig:examples}
\end{figure}

\xhdr{Qualitative Examples.}
We examine two qualitative examples of our Cross-Modal Attention (PM+DA\textsuperscript{*}+Aug.) model in unseen environments  (\figref{fig:examples}). The top example shows the agent successfully following the instruction and demonstrates the increased difficultly of VLN-CE (62 actions \vs 3 hops in VLN). Phrases like \myquote{turn left, and enter the hallway} present an additional challenge in VLN-CE as the agent must \texttt{turn-left} an unknown number of times until it sees the hallway.
The second example shows a failure of the agent -- 
it navigates towards the wrong windows and fails to first \myquote{pass the kitchen} -- stopping instead at the nearest couch. We also observe failures when the agent never sees the object(s) referred to by the instruction in the scene -- with a limited egocentric field-of-view, the agent must actively choose to observe the surrounding scene.

\csubsection{Examining the Impact of the Nav-Graph in VLN}
\label{sec:vlnce2vln}

To draw a direct comparison between the VLN and VLN-CE settings, we convert trajectories taken by our Cross-Modal Attention (PM+DA\textsuperscript{*}+Aug.) model in continuous environments to \navg trajectories (\SUPP{details in the supplement}) and then evaluate these paths on the VLN leaderboard.\footnote{Note that the VLN test set is not publicly available except through this leaderboard.} We emphasize that the point of this comparison is not to outperform existing approaches for VLN, but rather to highlight how important the \navg is to the performance of existing VLN systems by contrasting them with our model. Unlike the approaches shown, our model does not benefit from the \navg during training or inference.

As shown in \tabref{tab:test}, we find significant gaps between our model and prior work in the VLN setting. Despite having similar cross-modal attention architectures,  RCM \cite{wang2019reinforced} achieves an SPL of 0.38 in test environments while our model yields 0.21. Further, state-of-the-art on the test set is near 0.47 SPL, over 2x what we report. However, it is unclear if these gains could be realized on a real system given the strong assumptions set by the \navg. In contrast, our approach does not rely on external information and recent work has shown promising sim2real transferability for navigation agents trained in continuous simulations \cite{kadian_sim2realgap_2019}.

\xhdr{Caveats.} Making direct comparisons between drastically different settings is challenging, we note some caveats. Approximately 20\% of VLN trajectories are non-navigable in the continuous environments (and thus excluded in VLN-CE). By default, our models cannot succeed on these. Further, continuous VLN-CE paths  can translate poorly to nav-graph trajectories when traversing areas of the environment not well-covered by the sparse panoramas. Comparing VLN-CE val results in \tabref{tab:results} with the same in \tabref{tab:test} shows these effects account for a drop of approximately 10 SPL. Even compensating for this possible underestimation, \navg-based approaches still outperform our continuous models significantly.

\setlength{\tabcolsep}{.275em}
\begin{table}[t]
	\caption{Comparison on the VLN validation and test sets with existing models. Note there is a significant gap between techniques that leverage the oracle nav-graph at train and inference (top set) and our best method in continuous environments.
	}\vspace{-25pt}
	\label{tab:test}
	\begin{center}
	    \resizebox{\textwidth}{!}{
		\begin{tabular}{c>{\hspace{3pt}}l ccccs c ccccs c ccccs}
			\midrule
            & & \multicolumn{5}{c}{\scriptsize\textbf{Val-Seen (VLN)}} & & \multicolumn{5}{c}{\scriptsize\textbf{Val-Unseen (VLN)}} & & \multicolumn{5}{c}{\scriptsize\textbf{Test (VLN)}} \\
			\cmidrule{3-7} \cmidrule{9-13}
			\cmidrule{15-19}
			& \scriptsize Model &  \scriptsize\textbf{\texttt{TL}}~$\downarrow$ & \scriptsize\textbf{\texttt{NE}}~$\downarrow$ &
			 \scriptsize\textbf{\texttt{OS}}~$\uparrow$ & \scriptsize\textbf{\texttt{SR}}~$\uparrow$ & \scriptsize\textbf{\texttt{SPL}}~$\uparrow$ & & \scriptsize\textbf{\texttt{TL}}~$\downarrow$ & \scriptsize\textbf{\texttt{NE}}~$\downarrow$ &
			
			\scriptsize\textbf{\texttt{OS}}~$\uparrow$ & \scriptsize\textbf{\texttt{SR}}~$\uparrow$ & \scriptsize\textbf{\texttt{SPL}}~$\uparrow$& & \scriptsize\textbf{\texttt{TL}}~$\downarrow$ & \scriptsize\textbf{\texttt{NE}}~$\downarrow$ &  \scriptsize\textbf{\texttt{OS}}~$\uparrow$ & \scriptsize\textbf{\texttt{SR}}~$\uparrow$ & \scriptsize\textbf{\texttt{SPL}}~$\uparrow$\\
			\midrule
			\multirow{5}{*}[-0.5em]{\rotatebox{90}{VLN Task}}& VLN-Seq2Seq \cite{anderson2018vision}  & 11.33 & 6.01 & 0.52& 0.38 & -&   & 8.39 & 7.81 & 0.28 & 0.21 & - &   & 8.13 & 7.85 & 0.27 & 0.20 & 0.18\\
			& Self-Monitoring \cite{ma2019self}  & - & 3.18 & 0.77 & 0.68 & 0.58&   & - & 5.41 & 0.68 & 0.47 & 0.34 &   & 18.04 & 5.67 & 0.59 & 0.48 & 0.35\\
			& RCM \cite{wang2019reinforced}  & 10.65 & 3.53 & 0.75 & 0.66 & - &   & 11.46 & 6.09 & 0.50 & 0.42 & - &   & 11.97 & 6.12 & 0.495 & 0.43 & 0.38\\
			& Back-Translation \cite{tan2019learning}  & 10.1 & 4.71 & - & 0.55 & 0.53 &   & 9.37 & 5.49 & - & 0.46 & 0.43 &   & 11.7 & - & - & 0.51 & 0.47\\
			
			\cmidrule(l{0.25em}){2-19}
			\noalign{\vspace{0.25em}}
			
			& Cross-Modal {\scriptsize (PM+DA\textsuperscript{*}+Aug.)}  & 6.92 & 7.77 & 0.30 & 0.25 & 0.23&   & 7.42 & 8.17 & 0.28 & 0.22 & 0.20 &   & 9.47 & 8.55 & 0.32 & 0.24 & 0.21\\
			\bottomrule
		\end{tabular}}
	\end{center}
\end{table}

\csection{Discussion}
\label{sec:conc}

In this work, we explore the problem of following navigation instructions in continuous environments with low-level actions -- lifting many of the unrealistic assumptions in prior \navg-based settings. In models presented here, we took an approach where observations were mapped directly to low-level control in an end-to-end manner; however, exploring modular approaches is exciting future work. For instance, having the learned agent pass directives to a motion controller. Crucially, setting our VLN-CE task in continuous environments (rather than a \navg) provides the community with a testbed where these sort of integrative experiments studying the interface of high- and low-level control are possible.

\vspace{-0.11in}
\section*{Acknowledgements}
\vspace{-0.11in}

We thank Anand Koshy for his implementation of the dynamic time warping metric. The GT effort was supported in part by NSF, AFRL, DARPA, ONR YIPs, ARO PECASE, Amazon. The OSU effort was supported in part by DARPA. The views and conclusions contained herein are those of the authors and should not be interpreted as necessarily representing the official policies or endorsements, either expressed or implied, of the U.S. Government, or any sponsor.

\clearpage

{\small
\bibliographystyle{splncs04}
\bibliography{bib/strings,bib/main}
}
\clearpage
\section*{Supplementary}

\begin{figure}
    \centering
    \frame{\includegraphics[width=0.24\textwidth]{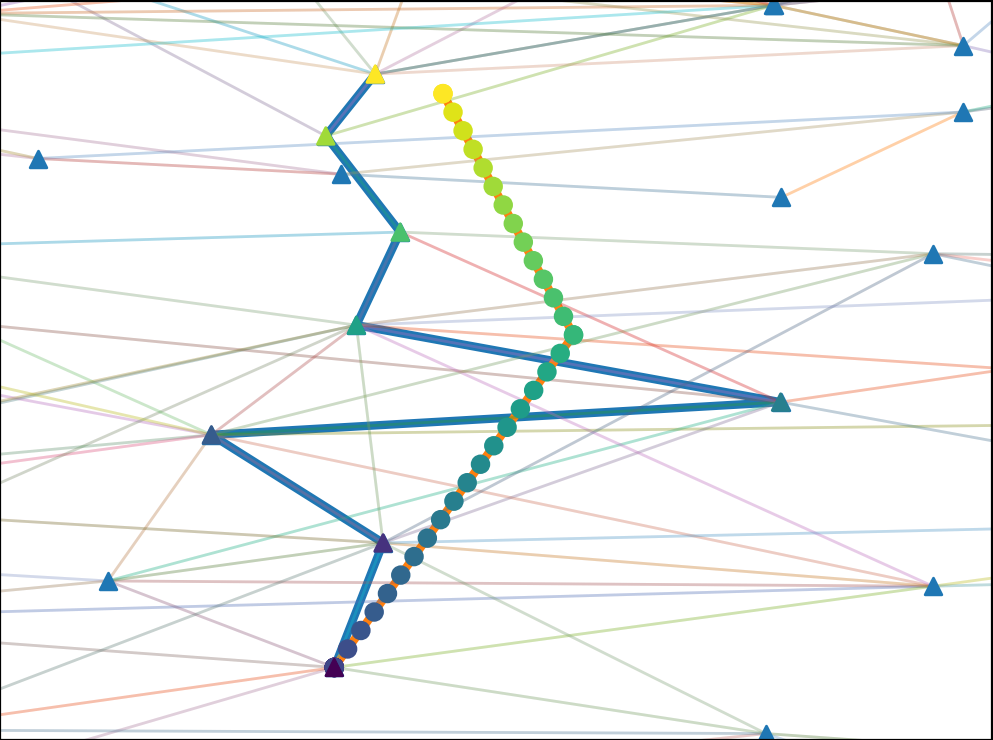}}
    \frame{\includegraphics[width=0.24\textwidth]{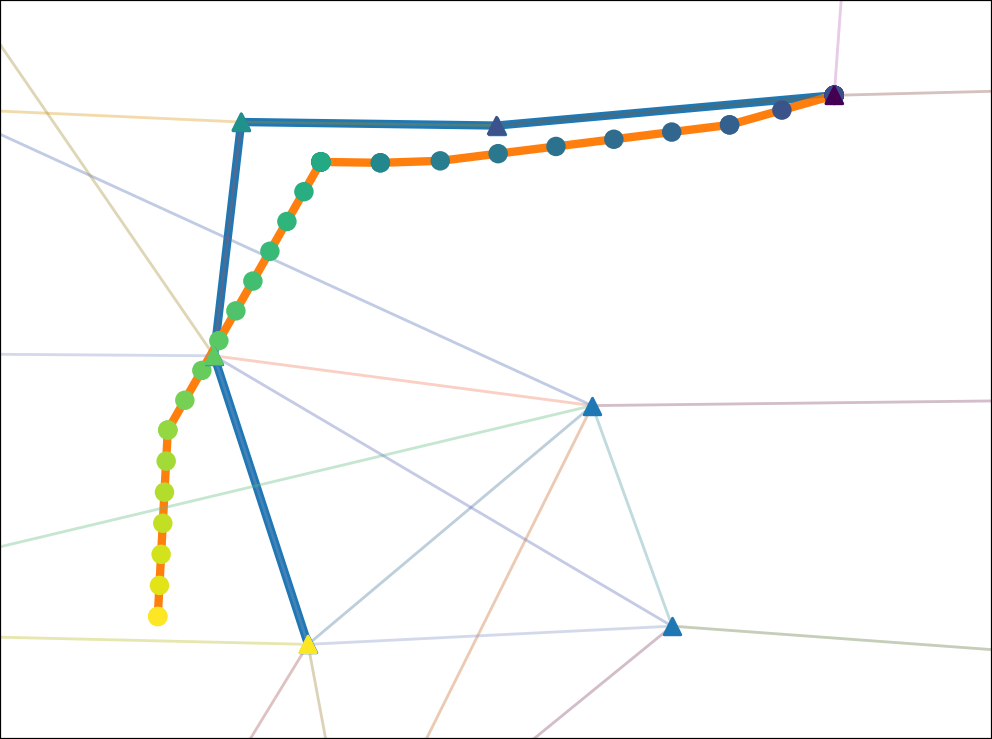}}
    \frame{\includegraphics[width=0.24\textwidth]{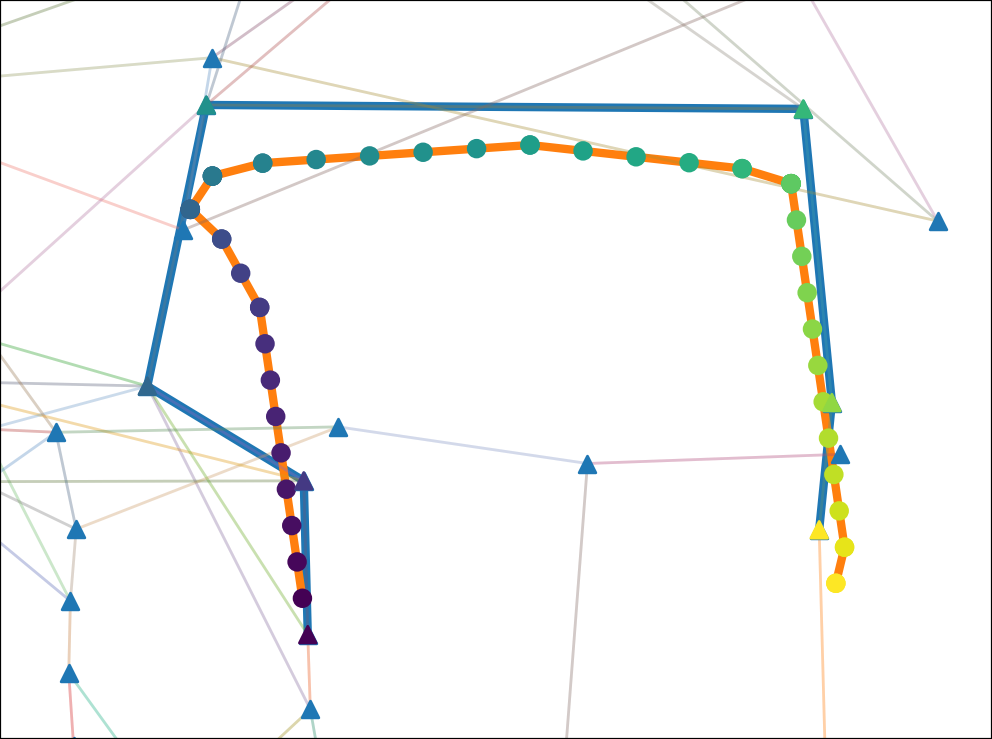}}
    \frame{\includegraphics[width=0.24\textwidth]{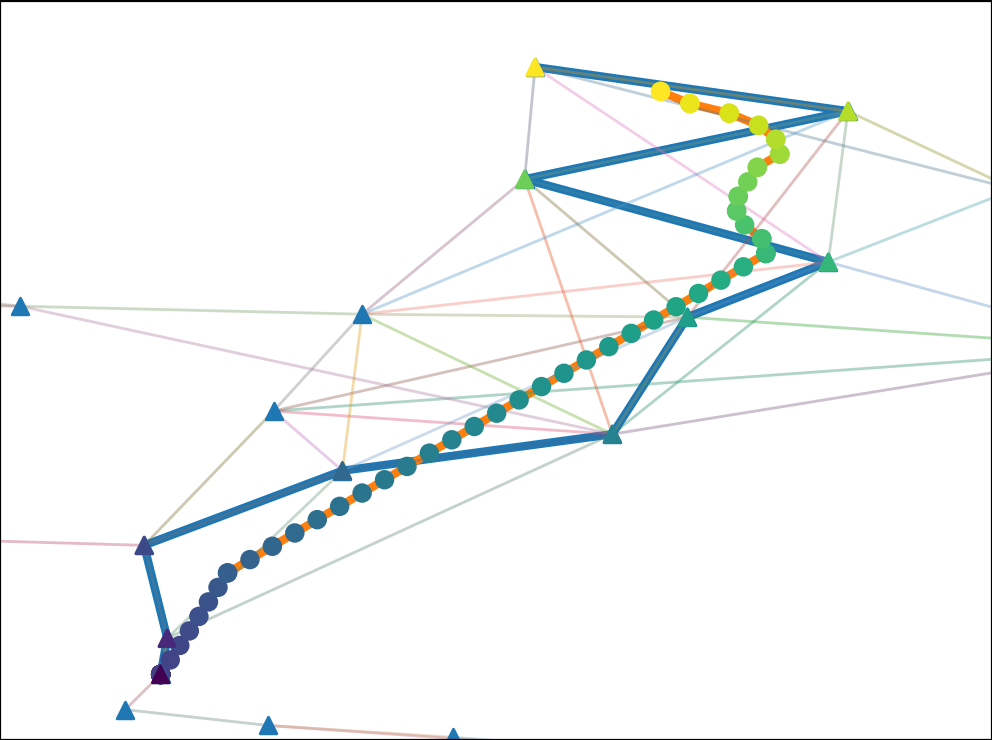}}
    \caption{Example conversions between VLN-CE paths (circles on orange line) and VLN nav-graph based trajectories (triangles on blue line). For both, darker markers are earlier in the trajectories. Other nav-graph nodes / edges are also shown. Note that the nav-graph is often a poor proxy for the 3D space with our agent paths in continuous space requiring zig-zag patterns in the nav-graph.}
    \label{fig:conversions}
    \vspace{-20pt}
\end{figure}
\csubsection{Converting VLN-CE Paths to Nav-Graph-based VLN}
As discussed in \secref{sec:vlnce2vln}, we convert paths from our agents in continuous environments to nav-graph trajectories for comparison with VLN. To do so, we apply a simple algorithm. 
Consider a trajectory in VLN-CE consisting of a sequence of positions $p_0, p_1, p_2 , \dots, p_T$. Note that the initial position $p_0$ aligns with the start node $v_s$ for this trajectory from the VLN nav-graph. The goal then is to find a path through the nav-graph from this node that follows the continuous path.

Starting from $v_s$, we iteratively snap to the nearest adjacent node by minimizing distance from the current position. More concretely, at the beginning of the sequence we set the current node $c$ to be $v_s$ and consider a `navigable set' $\mathcal{N}(c)$ consisting of all adjacent nodes to $c$ as well as $c$. We then compute the distance between every node in $\mathcal{N}(c)$ and the continuous environment path position $p_1$. Whichever node from the active set is nearest to $p_1$ becomes the new current node. This is repeated for $p_2,p_3,\dots, p_T$ with the current node (and thus navigable set) shifting to whichever node is nearest and within 1-step of the current node. 

\figref{fig:conversions} shows some of the resulting trajectories (triangles on thick blue line) as well as the underlying continuous path (dots on orange line). These images also show the nav-graph with thin multi-colored lines for edges and blue triangles for panoramic nodes. Notably, the nav-graph comes up short in representing the continuous paths. Often our agents will navigate through spaces not captured by nav-graph nodes; resulting in nav-graph trajectories that have high error or must oscillate to follow the continuous path.

\end{document}